\documentclass{article}
\usepackage{spconf,amsmath,amssymb,graphicx}
\usepackage{mathtools}
\usepackage{subfig}
\usepackage{siunitx}
\usepackage{bmpsize}
\usepackage[abs]{overpic}
\usepackage{adjustbox}
\usepackage[labelsep=period]{caption}
\renewcommand{\thetable}{\Roman{table}}

\usepackage{booktabs,array}
\usepackage{multirow}
\usepackage{balance}
 
\usepackage{pgfplots}
\usepackage{grffile}
\pgfplotsset{compat=newest}
\usetikzlibrary{plotmarks}
\usetikzlibrary{arrows.meta}
\usepgfplotslibrary{patchplots}
\usepackage{colortbl}%

\makeatletter
\let\NAT@parse\undefined
\makeatother

\usepackage[noadjust]{cite}

\usepackage[colorlinks=true,linkcolor=black, citecolor=black, urlcolor=black]{hyperref} 

\newcommand{\midsepremove}{\aboverulesep = 0mm \belowrulesep = 0mm}
\midsepremove
\newcommand{\midsepdefault}{\aboverulesep = 0.605mm \belowrulesep = 0.984mm}
\midsepdefault

\newcommand\blfootnote[1]{%
	\begingroup
	\renewcommand\thefootnote{}\footnote{#1}%
	\addtocounter{footnote}{-1}%
	\endgroup
}

\newcommand\figHeight{2.2in}

\definecolor{mycolor1}{rgb}{0.00000,0.44700,0.74100}%
\definecolor{mycolor3}{rgb}{0.92900,0.69400,0.12500}%
\definecolor{mycolor4}{rgb}{0.46667,0.67451,0.18824}%
\definecolor{mycolor2}{rgb}{0.85000,0.32500,0.09800}%

\definecolor{mycolor0}{rgb}{1,1,1}%
\definecolor{mycolor1}{rgb}{0.00000,0.45000,0.74000}%
\definecolor{mycolor2}{rgb}{0.85000,0.30000,0.10000}%
\definecolor{mycolor3}{rgb}{0.04706,0.47451,0.58039}%
\definecolor{mycolor4}{rgb}{0.00000,0.44700,0.74100}%
\definecolor{mycolor5}{rgb}{0.85000,0.32500,0.09800}%
\definecolor{mycolor6}{rgb}{0.00000,0.44700,0.74100}%
\definecolor{mycolor7}{rgb}{0.85000,0.32500,0.09800}%
\definecolor{mycolor8}{rgb}{0.92900,0.69400,0.12500}%
\definecolor{mycolor9}{rgb}{0.93000,0.69000,0.13000}%
\definecolor{mycolor10}{rgb}{0.49000,0.18000,0.56000}%
\definecolor{mycolor11}{rgb}{0.32000,0.49000,0.08000}%
\definecolor{mycolor12}{rgb}{0.85000,0.33000,0.10000}%
\definecolor{mycolor13}{rgb}{0.30000,0.42000,0.13000}%
\definecolor{mycolor14}{rgb}{0.64000,0.08000,0.18000}%
\definecolor{mycolor15}{rgb}{0.46667,0.67451,0.18824}%
\definecolor{mycolor16}{rgb}{0.49412,0.18431,0.55686}%

\newcommand{\myarrowdotted}[1][0.1pt]
{   \begin{tikzpicture}[overlay]
	\draw [->,>=stealth,line width=0.4mm,dashed,black] (-0.1, 0.1) -- (0.4, 0.1);
	\end{tikzpicture}
}

\newcommand{\myarrowsolid}[1][0.1pt]
{   \begin{tikzpicture}[overlay]
	\draw [->,>=stealth,line width=0.4mm,solid,black] (-0.1, 0.1) -- (0.4, 0.1);
	\end{tikzpicture}
}

\definecolor{blueelipse}{rgb}{0,0.2745,1}%
\definecolor{redelipse}{rgb}{0.784,0,0.1254}%
\definecolor{purpleelipse}{rgb}{0.5921,0.0470,0.8745}%
\newcommand{\myelipseblue}[1][0.1pt]
{   \begin{tikzpicture}[overlay]
	\draw [line width=0.5mm,blueelipse](0, 0.1) ellipse (1mm and 1mm);
	\end{tikzpicture}
}
\newcommand{\myelipsered}[1][0.1pt]
{   \begin{tikzpicture}[overlay]
	\draw [line width=0.5mm,redelipse](0, 0.1) ellipse (1mm and 1mm);
	\end{tikzpicture}
}
\newcommand{\myelipsepurple}[1][0.1pt]
{   \begin{tikzpicture}[overlay]
	\draw [line width=0.5mm,purpleelipse](0, 0.1) ellipse (1mm and 1mm);
	\end{tikzpicture}
}

\makeatletter
\renewcommand\footnotesize{%
	\@setfontsize\footnotesize\@ixpt{11.4}%
	\abovedisplayskip 8\p@ \@plus2\p@ \@minus4\p@
	\abovedisplayshortskip \z@ \@plus\p@
	\belowdisplayshortskip 4\p@ \@plus2\p@ \@minus2\p@
	\def\@listi{\leftmargin\leftmargini
		\topsep 4\p@ \@plus2\p@ \@minus2\p@
		\parsep 2\p@ \@plus\p@ \@minus\p@
		\itemsep \parsep}%
	\belowdisplayskip \abovedisplayskip
}
\makeatother
\title{Uncertainty-based Biological Age Estimation of Brain MRI scans}

\name{Karim Armanious\textsuperscript{1,2}, Sherif Abdulatif\textsuperscript{~1}, Wenbin Shi\textsuperscript{1}, 
	Tobias Hepp\textsuperscript{3}, Sergios Gatidis\textsuperscript{2}, Bin Yang\textsuperscript{1}}
\address{\textsuperscript{1}University~of~Stuttgart,~Institute~of~Signal~Processing~and~System~Theory,~Stuttgart,~Germany\\
	\textsuperscript{2}University~of~T\"ubingen,~Department~of~Radiology,~T\"ubingen,~Germany\\
	\textsuperscript{3}Max Planck Institute for Intelligent Systems,~Empirical Inference Department,~T\"ubingen,~Germany
	\thanks{An extended version of this paper is available in \url{https://arxiv.org/abs/2009.10765} \cite{Final3}.}
	\thanks{The first two authors equally contributed to this work.}
}
\begin{document}
	%
	\maketitle
	\begin{abstract}
		
		Age is an essential factor in modern diagnostic procedures. However, assessment of the true biological age (BA) remains a daunting task due to the lack of reference ground-truth labels. Current BA estimation approaches are either restricted to skeletal images or rely on non-imaging modalities that yield a whole-body BA assessment. However, various organ systems may exhibit different aging characteristics due to lifestyle and genetic factors. In this initial study, we propose a new framework for organ-specific BA estimation utilizing 3D magnetic resonance image (MRI) scans. As a first step, this framework predicts the chronological age (CA) together with the corresponding patient-dependent aleatoric uncertainty. An iterative training algorithm is then utilized to segregate atypical aging patients from the given population based on the predicted uncertainty scores. In this manner, we hypothesize that training a new model on the remaining population should approximate the true BA behavior. We apply the proposed methodology on a brain MRI dataset containing healthy individuals as well as Alzheimer’s patients. We demonstrate the correlation between the predicted BAs and the expected cognitive deterioration in Alzheimer’s patients.

	\end{abstract}
	\vspace{-1mm}
	\section{Introduction}
	\vspace{-1mm}
	\label{sec:intro}
	
	Age is one of the most important parameters describing individuals in a medical context. However, age-related biological phenotypes can deviate significantly between individuals within the same age group. These observations have motivated the concept of biological age (BA) in contrast to chronological age (CA) \cite{RN6}. CA is described as the amount of time since the birth of an individual. On the other hand, BA can be described as a measure for the extent of genetic, metabolic and functional changes in an individual that occur during the process of aging. Thus, BA can be considered as an extension to the traditional concept of CA in addition to any organ-specific accelerated or delayed aging characteristics \cite{RN7,RN8,RN10}. Despite this relatively imprecise definition, the potential impact of the concept of BA is easily conceivable. It is a common practice for clinicians to assess the overall condition of patients relative to their respective age group and incorporate this impression into their medical decisions. However, these personal estimates are subjective and not easily quantifiable.	
	
	A large body of research attempts the quantification of BA using non-imaging data \cite{E5,E6,E7}. More specifically, age-dependent variables such as genetic \cite{RN6,RN7}, cellular \cite{RN8}, phenotypic and epidemiological data \cite{RN10,RN9}, blood biomarkers \cite{E1,E2} and physical activity \cite{E3} have been used as indicators for the BA. The majority of these approaches utilize cohort datasets for the prediction of the mortality risk \cite{E8,E9,E10}. Other methods incorporate the CA as ground-truth labels and examine the relation between the predicted ages and other health indicators for assessing the BA \cite{E11,E12}. Nonetheless, the above non-imaging approaches lead to a whole-body assessment of the BA. In that sense, they are not capable of recognizing the differences in aging characteristics between individual organ systems. In this context, medical imaging may potentially provide significant information allowing for non-invasive estimation of organ-specific BA.

	\begin{figure*}[!t]
		\centering
		\includegraphics[width=0.98\textwidth]{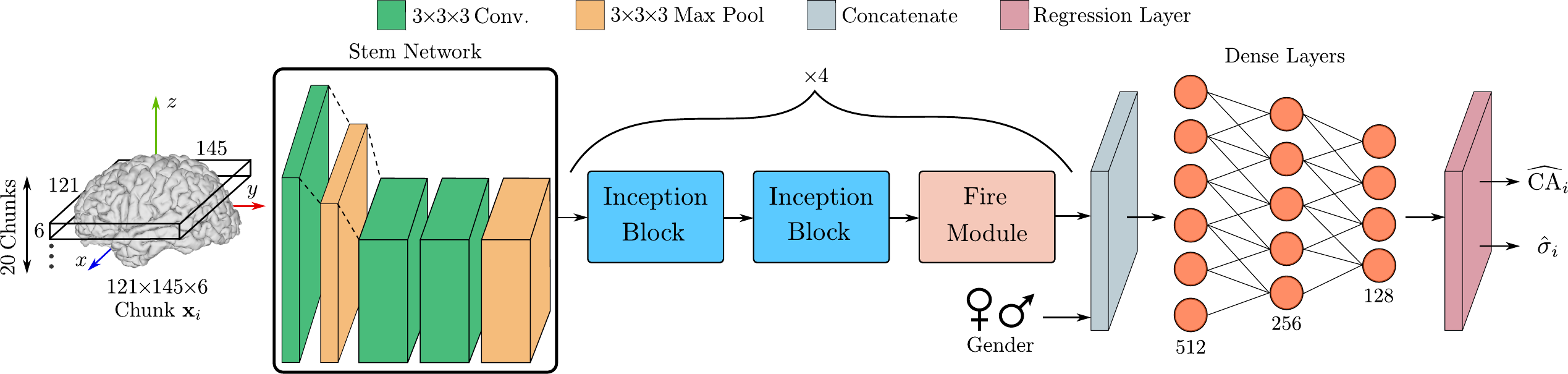}
		\caption{An overview of the Age-Net architecture utilized for the iterative BA estimation framework. The network takes as input an MRI chunk $\mathbf{x}_i$ of segmented brain gray matter together with the gender label of the patient. The outputs of this network represent the predicted mean age $\smash{\widehat{\textrm{CA}}_i}$ and the corresponding aleatoric uncertainty prediction $\hat{\sigma}_i$ of the $i^{\textrm{th}}$ chunk. \label{fig:architecture}}
		\vspace{-5mm}
	\end{figure*}
	
	The use of medical imaging for age estimation in a clinical setting is limited mainly to skeletal age estimation in infants and adolescents using conventional radiography and MRI \cite{RN14,Final1,x17,x18-3}. Beyond skeletal age estimation, first studies have introduced the concept of brain age based on changes in brain morphology and provided evidence for associations between premature brain aging and cognitive functions \cite{RN15,Final2}. Similarly, Alzheimer’s disease was found to correlate directly with abnormal brain aging \cite{E16}. In recent years, the use of convolutional neural networks (CNNs) has become more prevalent due to strong results in a multitude of tasks \cite{ipa,x24,Extra2,x25,Final6,rad0,rad2}. For instance, a deep CNN has been utilized for prediction of brain age using 2D T1-weighed MR images \cite{x11}. Recent advances have attempted the use of relatively shallow 3D CNN architectures to incorporate the spatial information between slices in the brain age estimation procedure \cite{x27,x15}.
	
	As stated clearly in the recent survey on this topic \cite[p.~11]{E13}, all deep learning (DL) approaches, whether imaging or non-imaging based, relies on the CA as ground-truth labels for BA prediction \cite{E14}. Thus, the predicted ages can not be used to assess the true aging behavior of the test subjects. To the best of our knowledge, the problem of defining ground-truth BA labels is still not possible and remains an open research question. From another perspective, all above DL approaches also provide point estimates which are incapable of assessing the uncertainty in the network predictions.	However, as medical applications may generally influence critical therapeutic decisions, recent years have witnessed the utilization of uncertainty prediction techniques in the medical domain  \cite{UN1,UN2,UN3}. On the whole, two main approaches are adopted for uncertainty predictions. Namely, aleatoric uncertainty reflects the noise inherent within the input dataset \cite{Final7} and epistemic approximations which accounts for the model uncertainty \cite{Final5}.
	
	
	The purpose of this study is to bridge the gap between chronological and biological age estimation. This is achieved by incorporating a CA estimation framework, named ``Age-Net'', as the baseline for the age regression task \cite{Extra}. Furthermore, we introduce a novel iterative training strategy for approximating organ-specific BA labels. This framework attempts to identify, and subsequently segregate, outliers who exhibit atypical-aging characteristics. To accomplish this, the CNN-based Age-Net is expanded to additionally estimate the aleatoric uncertainty. This was chosen to model the patient-dependent uncertainty of the input MRI scans, which is later exploited for the detection of outlier patients (BA $\not \approx$ CA). To validate the accuracy of the utilized training approach, we apply the proposed methodology on a brain MRI dataset containing both healthy and Alzheimer's affected patients. Subsequently, we quantify the amount of Alzheimer's patients detected as atypically-aging patients by the iterative strategy. Interested readers can refer to \cite{Final3} for an extended analysis of the proposed iterative training algorithm.

	\vspace{-2.5mm}
	\section{Architecture}
	\vspace{-1.5mm}
	In this section, we present the Age-Net architecture, which was originally introduced for organ-based CA regression task. To extend this framework for BA estimation, we expand the architecture to additionally predict the aleatoric uncertainty. This is later incorporated by the iterative training framework to segregate atypically aging patients. An outline of the utilized architecture is depicted in Fig.~\ref{fig:architecture}.
	
	\vspace{-3mm}
	\subsection{Age-Net}
	\vspace{-1mm}
	The Age-Net architecture for CA estimation is constructed out of a hybrid combination of inception v1 \cite{x5} and Squeeze-Net \cite{x4} architectures. These modules assist in reducing the total number of trainable parameters while enhancing the representation capacity. The final architecture is composed of an initial stem network followed by four modules concatenated together in an end-to-end manner. Each module consists of two inception blocks, followed by a single fire module. Finally, three dense layers combine additional gender information before a final regression layer outputs the predicted age. Based on the initial comparative analysis presented in \cite{Extra}, a chunk data-feeding strategy is adopted where the input MRI volumes are divided into smaller 3D chunks before being fed as inputs to the network. 
	\vspace{-3mm}
	\subsection{Heteroscedastic Aleatoric Uncertainty}
	\vspace{-1mm}
	In order to detect patients with atypical aging characteristics, it is beneficial to identify how certain the regression network is with regards to the predicted ages. To this end, we employ a Gaussian regression process which estimates the aleatoric uncertainty, as previously proposed in \cite{Final7,Final5}. This can be used to either model the uncertainty inherent in each input MRI chunk (i.e. heteroscedastic uncertainty) or to represent the uncertainty of the entire MRI dataset (i.e. homoscedastic uncertainty). We employ the heteroscedastic variant for patient-dependent uncertainty predictions. As such, the last dense layer of the Age-Net architecture is modified to have two output nodes representing the mean predicted age $\smash{\widehat{\textrm{CA}}_i}$ and the corresponding uncertainty prediction $\hat{\sigma}_i$ for each input chunk $i$. For training this framework, a Gaussian likelihood distribution is assumed, leading to the following maximum likelihood (ML) estimation \cite{Final4}:   
	\begin{equation}
		\begin{split}
		\mathcal{L}_{\textrm{ML}}  & =  \max_\omega \log p(\mathcal{D}|\omega)\\
		& = \frac{1}{N} \sum_{i=1}^{N} \frac{1}{2 \hat{\sigma}_i^2} \left( \textrm{CA}_i - \widehat{\textrm{CA}}_i \right)^2 + \frac{1}{2} \log \hat{\sigma}_i^2
		\end{split}
		\vspace{-1mm}
	\end{equation}
	where $p(\mathcal{D}|\omega)$ is the likelihood function over the given training dataset of MRI chunks and their corresponding CA labels $\mathcal{D} = \left\{ \left(\mathbf{x}_i, \textrm{CA}_i\right) \right\}_{i=1}^N$. The total number of training chunks is represented by $N$, whereas $\omega$ are the trainable weights. 
	\vspace{-2mm}
	\section{BA Estimation Framework}
	\vspace{-1mm}
	\begin{figure*}[t]
		\centering
		\includegraphics[width=0.95\textwidth]{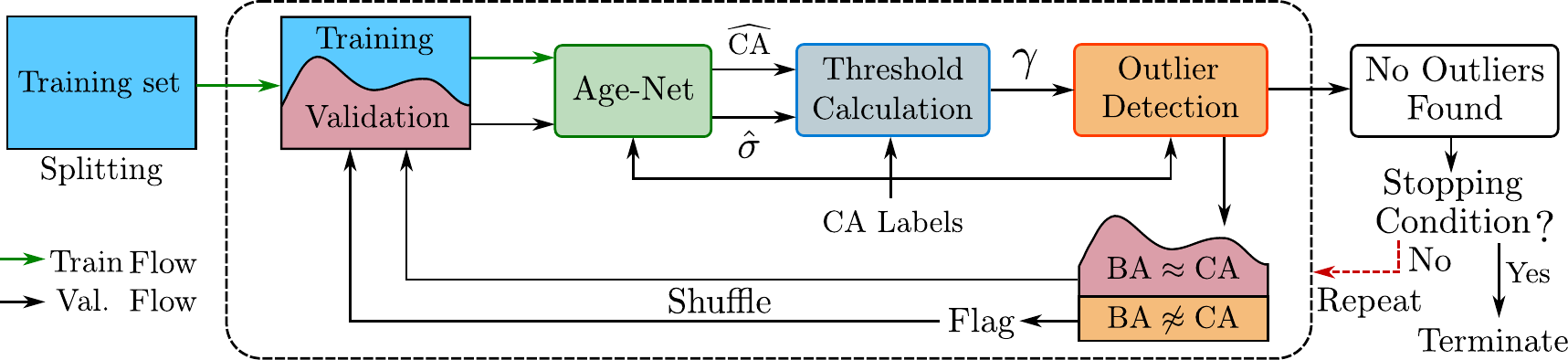}
		\caption{A detailed flow chart of the proposed iterative training strategy for the extraction of BA labels. \label{fig:iterative_cleaning}}
		\vspace{-4.5mm}
	\end{figure*} 
	The main challenge for MRI-based BA estimation is the lack of ground-truth labels. To resolve this challenge, we propose an iterative training strategy, illustrated in Fig.~\ref{fig:iterative_cleaning}, to approximate BA labels.
	\vspace{-3.5mm}
	\subsection{Iterative Training Strategy}
	\vspace{-1.5mm}
	The first step in each iteration of the proposed strategy is to shuffle and split the training data into two subsets. The first subset is used to train the Age-Net architecture. The trained model is then validated on the second subset and the estimated ages $( \, \smash{\widehat{\textrm{CA}}} \,)$ together with their uncertainty predictions $( \hat{\sigma})$ are used to calculate a patient-dependent threshold $\gamma$. This threshold is then utilized for the detection of outliers who exhibit atypical-aging characteristics in the validation subset. The process of threshold calculation and outlier detection is explained in the next subsection. The identified patients are then flagged as outliers. A new iteration would then be repeated, starting with merging all validation samples (including the outliers) with the training subset, reshuffling and repeating the process by training the Age-Net framework from scratch. An empirical stopping condition is enabled, which states that three consecutive data-cleaning iterations trained with different initializations must yield no new outliers before the iterative strategy can be terminated. Upon termination of the training algorithm, all patients who were flagged as outliers in more than one iteration are removed from the training dataset. This serves to assert that no typically-aging patient is wrongfully detected as an outlier. Finally, an Age-Net architecture is trained on the cleaned dataset where the CA labels should correspond approximately to the true BA labels ($\textrm{BA} \approx \textrm{CA}$).
	\vspace{-3mm}
	\subsection{Outliers Detection}
	In this work, we utilize a chunk data feeding strategy where each input MR volume is divided into $K$ smaller chunks before being fed as inputs to the Age-Net. This process yields two outputs for each input chunk $\mathbf{x}_i$: the predicted age score $\smash{\widehat{\textrm{CA}}_i}$ and the corresponding aleatoric uncertainty prediction $\hat{\sigma}_i$. Outlier detection is initiated by first calculating a consolidated $\smash{\widehat{\textrm{CA}}}$ and uncertainty estimates for each patient in the validation dataset. This is achieved by averaging out the predicted ages and uncertainties for each chunk $i$ in the MR volume of patient $n$ as: 
	\begin{equation}
	\left\{\widehat{\textrm{CA}}_n \, , \, \hat{\sigma}_n  \right\}  = \frac{1}{K} \sum_{i=1}^{K} \left\{ \widehat{\textrm{CA}}_{n,i} \, , \, \hat{\sigma}_{n,i} \right\} 
	\end{equation}
	This process is then repeated for all patients. For each patient, we compare the predicated age deviation ($\textrm{D}_n$) against a patient-dependent threshold ($\gamma_{n}$), defined as:
	\begin{equation}
	\textrm{D}_n = \left| \widehat{\textrm{CA}}_n - \textrm{CA}_n\right| \, \, \, \, \, \,  \,,   \,\, \, \, \, \, \,	\gamma_{n} = R \cdot \hat{\sigma}_n
	\vspace{-3mm}
	\end{equation}
	where $R$ is a pre-defined constant value. The $n^{\textrm{th}}$ patient is flagged as an outlier only if the age deviation exceeds the corresponding threshold value as represented by ${\textrm{D}}_n > \gamma_{n}$.
	
	Assuming a normal distribution for the chunk predictions, the arbitrary constant $R$ was set to $1.96$ to reflect the $95$\% confidence interval of the mean predicted age of each patient. After each iteration, the training and validation datasets are reshuffled, and a new iteration would commence until the stopping condition is reached. Upon termination of the algorithm, all patients detected as outliers are removed from the training dataset only if they were flagged in more than one iteration. The final framework is then trained on this dataset containing only patients with typical-aging characteristics.
	\vspace{-2mm}
	\section{Datasets and Experiments}
	\vspace{-1mm}
	We investigate the performance of the introduced training strategy on a subset from the OASIS-3 brain dataset \cite{Extra6}. This dataset encompasses T1-weighted MR scans from anonymized cognitively healthy individuals as well as patients suffering from dementia due to Alzheimer's disease. The degree of cognitive deterioration in Alzheimer's patients is indicated by the clinical dementia rating (CDR) which distinguishes between questionable, mild and moderate dementia by the CDR scores of 0.5, 1 and 2, respectively \cite{Extra7}. In total, we utilize a subset of 1240 MRI scans from 950 patients in the age range of 48-97 years. We allocate 565 MR scans from 405 healthy patients and 185 scans from 165 Alzheimer's patients for training the proposed framework. The remaining 490 scans from 380 patients (270: healthy, 110: Alzheimer's) are assigned as the test set. The same pre-processing pipeline described previously in \cite{Extra} was also applied with MR chunks of matrix size $121 \times 145 \times 6$ being fed to the framework as inputs.

	In previous studies, it has been reported that Alzheimer's disease correlates directly with abnormal brain characteristics, particularly accelerated aging \cite{E16}. We apply this observation as an attempt to evaluate the capability of the iterative strategy in detecting atypically-aging individuals. More specifically, we count the number of patients flagged as outliers by the proposed training strategy. Further, we analyze the percentage of cognitively healthy individuals (CDR = 0), and Alzheimer's patients (CDR = 0.5, 1, 2) detected as outliers with respect to their corresponding populations in the training dataset. We hypothesize that the proposed training strategy should be capable of accurately detecting patients with mild and moderate dementia as these patients exhibit the most pronounced atypical-aging characteristics. 
	
	\begin{figure}[!t]
		\resizebox{0.98\columnwidth}{!}{
%
%
%
\begin{tikzpicture}

\begin{axis}[%
width=4.52in,
height=\figHeight,
at={(0in,0in)},
scale only axis,
xmin=0,
xmax=19,
xlabel style={font=\color{white!15!black}},
xlabel={\large Iteration},
ymin=0,
ymax=105,
ylabel style={font=\color{white!15!black}},
ylabel={\large Cumulative Outliers [\%]},
ytick = {0,20,40,60,80,100},
axis background/.style={fill=white},
legend style={at={(0.015,0.565)}, anchor=south west, legend cell align=left, align=left, draw=white!15!black}
]
\addplot [color=mycolor4, line width=2.0pt, mark size=1.5pt, mark=*, mark options={solid, fill=mycolor4, mycolor4}]
  table[row sep=crcr]{%
1	3.70370370370370\\
2	11.1111111111111\\
3	15.3086419753086\\
4	19.7530864197531\\
5	20.7407407407407\\
6	21.7283950617284\\
7	23.4567901234568\\
8	24.1975308641975\\
9	24.6913580246914\\
10	25.9259259259259\\
11	27.4074074074074\\
12	28.3950617283951\\
13	28.6419753086420\\
14	29.3827160493827\\
15	30.1234567901235\\
16	31.8518518518519\\
17	31.8518518518519\\
18	31.8518518518519\\
};
\label{lab:cdr_0}
\addlegendentry{Normal}


\addplot [color=mycolor5, line width=2.0pt, mark size=1.5pt, mark=*, mark options={solid, fill=mycolor5, mycolor5}]
  table[row sep=crcr]{%
1	12.7272727272727\\
2	20.9090909090909\\
3	30.9090909090909\\
4	38.1818181818182\\
5	38.1818181818182\\
6	40\\
7	44.5454545454546\\
8	49.0909090909091\\
9   51.8181818181818\\
10	53.6363636363636\\
11	53.6363636363636\\
12	55.4545454545455\\
13	58.1818181818182\\
14	61.8181818181818\\
15	62.7272727272727\\
16	64.5454545454546\\
17	64.5454545454546\\
18	64.5454545454546\\
};
\label{lab:cdr_half}
\addlegendentry{CDR $0.5$}

\addplot [color=mycolor15, line width=2.0pt, mark size=1.5pt, mark=*, mark options={solid, fill=mycolor15, mycolor15}]
  table[row sep=crcr]{%
1	8.16326530612245\\
2	36.7346938775510\\
3	42.8571428571429\\
4	46.9387755102041\\
5	48.9795918367347\\
6	55.1020408163265\\
7	57.1428571428571\\
8	63.2653061224490\\
9	65.3061224489796\\
10	67.3469387755102\\
11	69.3877551020408\\
12	69.3877551020408\\
13	69.3877551020408\\
14	71.4285714285714\\
15	71.4285714285714\\
16	73.469387755102\\
17	73.469387755102\\
18	73.469387755102\\
};
\label{lab:cdr_1}
\addlegendentry{CDR $1$}

\addplot [color=mycolor16, line width=2.0pt, mark size=1.5pt, mark=*, mark options={solid, fill=mycolor16, mycolor16}]
  table[row sep=crcr]{%
1	16.6666666666667\\
2	66.6666666666667\\
3	83.3333333333333\\
4	83.3333333333333\\
5	83.3333333333333\\
6	83.3333333333333\\
7	100\\
8	100\\
9	100\\
10	100\\
11	100\\
12	100\\
13	100\\
14	100\\
15	100\\
16	100\\
17	100\\
18	100\\
};
\label{lab:cdr_2}
\addlegendentry{CDR $2$}
\node[right, align=left]
at (axis cs:14,38) {$129/405 = 32$\%};
\node[right, align=left]
at (axis cs:14.3,55) {$71/110 = 65$\%};
\node[right, align=left]
at (axis cs:14.6,79) {$36/49 = 73$\%};
\node[right, align=left]
at (axis cs:14.9,93) {$6/6 = 100$\%};
\end{axis}
\end{tikzpicture}%
}
		\vspace{-2mm}\caption{The cumulative outliers detected during the proposed iterative strategy with respect to the total number of patients in the training dataset from the corresponding CDR levels. \label{fig:data_trial}}
		\vspace{-5mm}
	\end{figure}
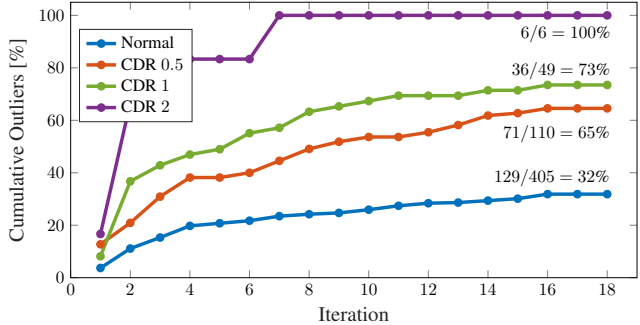
	Additionally, we compare the predicted BAs, after applying the iterative training algorithm, against an Age-Net trained conventionally by using the CA as ground-truth labels. We illustrate and subsequently analyze the distributions of the resultant age estimates of the two frameworks. This was conducted separately for both cognitively healthy and Alzheimer's patients.
	\vspace{-2.5mm}
	\section{Results and Discussion}
	\vspace{-1.5mm}
	The first step towards analyzing the proposed iterative training strategy is to examine the detected outliers across consecutive iterations. As depicted in Fig.~\ref{fig:data_trial}, a total of 18 training iterations were conducted before termination. This is due to satisfying the pre-defined stopping condition with no outliers detected in three successive iterations. Upon examining the patients with moderate dementia (CDR = 2, \ref{lab:cdr_2}), it is observed that the proposed strategy detects all aforementioned patients after 7 iterations. For Alzheimer's patients with mild dementia (\ref{lab:cdr_1}), 36 out of 49 patients were flagged as outliers in 16 iterations amounting to a total of 73\% of the CDR 1 training population. For questionable dementia (CDR = 0.5, \ref{lab:cdr_half}), 65\% of this population were detected as outliers in 16 iterations. Conversely, for cognitively healthy individuals (\ref{lab:cdr_0}), a substantially smaller percentage of patients (32\%) were flagged as outliers. Compared to the number of outliers from Alzheimer's patients, it is realistic for cognitively healthy individuals to less frequently exhibit atypically-aging characteristics. The above findings indicate that this algorithm is capable of detecting atypical-aging characteristics, whether in Alzheimer's patients or healthy individuals. 
	
	\begin{figure}
		\captionsetup[subfigure]{oneside,margin={1.cm,0cm}}
		\subfloat[Cogneitvely healthy patients (CDR = 0).\label{fig:deviations_healthy}]{\resizebox{0.98\columnwidth}{!}{\input{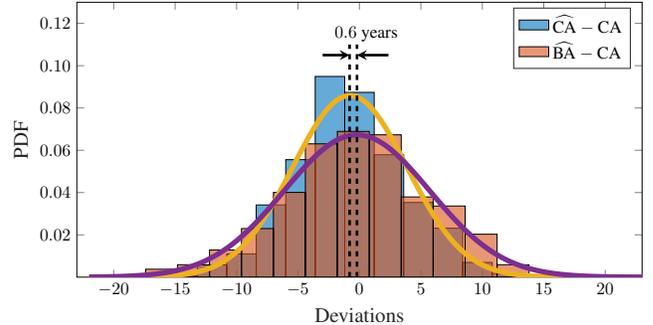}}}
		\vspace{-2mm}
		\subfloat[Alzheimer's patients (CDR $>$ 0).\label{fig:deviations_alz}]{\resizebox{0.98\columnwidth}{!}{\input{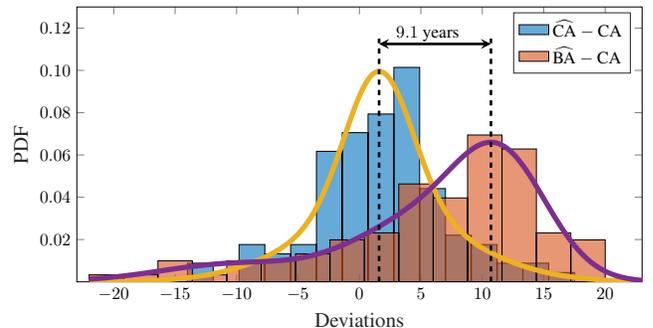}}}
		\vspace{-2mm}\caption{The PDF of the deviations between the estimated ages and the ground-truth CA labels. The depicted lines (\ref{lab:healthy_ca}) and (\ref{lab:healthy_ba}) represents the best-fit distribution for the CA and BA networks, respectively. \label{fig:deviations_all}}
		\vspace{-6mm}
	\end{figure}
	
	We also compare the predicted ages of the proposed BA estimation framework against those from a conventionally trained Age-Net with CA labels as ground-truth. The probability distribution functions (PDFs) of the deviations between the predicted ages and ground-truth CA labels for both frameworks are presented in Fig.~\ref{fig:deviations_all}. For the cognitively healthy population, minor perceivable differences can be observed in the distributions of both frameworks. This is illustrated in Fig.~\ref{fig:deviations_healthy} with both of them adopting a normal distribution with a mean-deviation shift of only 0.6 years. However, in Fig.~\ref{fig:deviations_alz} the majority of the predicted BAs (\ref{lab:alz_ba}) exhibit an over-aging of approximately 9.1 years compared to the CA framework (\ref{lab:alz_ca}). We hypothesize that this reflects the capability of the proposed framework in recognizing true BA behavior, unlike conventional CA estimation frameworks.
	
	Despite the above results, this initial study is not without limitations. In the future, we plan to expand upon the current framework with epistemic approximations to represent the model uncertainty by incorporating Bayesian neural networks and variational inference. This can be used to further refine the outlier detection procedure. Moreover, clinical assessments by radiologists are necessary for the validation of the introduced framework. More specifically, the generalization potential of this framework should be investigated with regards to the detection of different disorders, e.g. brain tumors and lesions. Another key question is the applicability of the BA framework on whole-body MRI dataset for age estimation of various organ systems.

	\vspace{-3.5mm}
	\section{Conclusion}
	\vspace{-6mm}
	\blfootnote{This work is funded by the DFG: 42821930/SPP 2177.}
	
	In this study, we propose an MRI-based framework for organ-specific BA estimation. As an initial step, we apply this framework on brain MRI scans. This framework encompasses a convolutional architecture for CA estimation as well as an iterative training algorithm. This algorithm exploits the heteroscedastic aleatoric uncertainty predictions for the detection of outlier patients exhibiting atypical-aging characteristics. By segregating the detected patients, this leads to the creation of a dataset where the available CA labels approximate the BA behavior. Upon validating the proposed BA framework on an Alzheimer's dataset, the majority of patients with mild or moderate dementia were detected as outliers. Moreover, the framework was found to be effective in detecting accelerated aging in Alzheimer's patients in comparison to conventional CA estimation. 
	%

	
	\bibliographystyle{IEEEbib}

\begin{thebibliography}{10}
		\setlength\parskip{0pt}
		\fontsize{8.19pt}{8.19pt}\selectfont
		\providecommand{\url}[1]{#1}
		\csname url@samestyle\endcsname
		\providecommand{\newblock}{\relax}
		\providecommand{\bibinfo}[2]{#2}
		\providecommand{\BIBentrySTDinterwordspacing}{\spaceskip=5pt\relax}
		\providecommand{\BIBentryALTinterwordstretchfactor}{4}
		\providecommand{\BIBentryALTinterwordspacing}{\spaceskip=\fontdimen2\font plus
			\BIBentryALTinterwordstretchfactor\fontdimen3\font minus
			\fontdimen4\font\relax}
		\providecommand{\BIBforeignlanguage}[2]{{%
				\expandafter\ifx\csname l@#1\endcsname\relax
				\typeout{** WARNING: IEEEtran.bst: No hyphenation pattern has been}%
				\typeout{** loaded for the language `#1'. Using the pattern for}%
				\typeout{** the default language instead.}%
				\else
				\language=\csname l@#1\endcsname
				\fi
				#2}}
		\providecommand{\BIBdecl}{\relax}
		\BIBdecl
		
	\bibitem{Final3}
	K.~Armanious et~al.,
	\newblock ``{Age-Net: An MRI-based iterative framework for brain biological age
		estimation},'' \url{https://arxiv.org/abs/2009.10765}, 2020,
	\newblock arXiv preprint.
	
	\bibitem{RN6}
	L.~Jia, W.~Zhang, and X.~Chen,
	\newblock ``Common methods of biological age estimation,''
	\newblock {\em Clinical Interventions in Aging}, vol. 12, pp. 759--772, 2017.
	
	\bibitem{RN7}
	B.~H. Chen et~al.,
	\newblock ``DNA methylation-based measures of biological age: meta-analysis
	predicting time to death,''
	\newblock {\em Aging}, vol. 8, no. 9, pp. 1844--1865, 2016.
	
	\bibitem{RN8}
	V.~Ignjatovic et~al.,
	\newblock ``Age-related differences in plasma proteins: how plasma proteins
	change from neonates to adults,''
	\newblock {\em PLoS One}, vol. 6, no. 2, 2011.
	
	\bibitem{RN10}
	E.~Nakamura and K.~Miyao,
	\newblock ``A method for identifying biomarkers of aging and constructing an
	index of biological age in humans,''
	\newblock {\em The journals of gerontology. Series A, Biological sciences and
		medical sciences}, vol. 62, no. 10, pp. 1096--1105, 2007.
	
	\bibitem{E5}
	P. Klemera and S. Doubal,
	\newblock ``A new approach to the concept and computation of biological age,''
	\newblock {\em Mechanisms of ageing and development}, vol. 127, pp. 240--248, 2006.
	
	\bibitem{E6}
	P. Fedichev et~al.,
	\newblock ``Extracting biological age from biomedical data via deep learning:
	Too much of a good thing?,''
	\newblock {\em Scientific Reports}, vol. 8, 2018.
	
	\bibitem{E7}
	E. Putin et~al.,
	\newblock ``Deep biomarkers of human aging: Application of deep neural networks
	to biomarker development,''
	\newblock {\em Aging}, vol. 8, no. 5, pp. 1021--1033, 2016.
	
	\bibitem{RN9}
	J.~Jylhava, N.~L. Pedersen, and S.~Hagg,
	\newblock ``{Biological age predictors},''
	\newblock {\em EBioMedicine}, vol. 21, pp. 29--36, 2017.
	
	\bibitem{E1}
	S.~A. {Rahman} and D.~A. {Adjeroh},
	\newblock ``Centroid of age neighborhoods: A new approach to estimate
	biological age,''
	\newblock {\em IEEE Journal of Biomedical and Health Informatics}, vol. 24, no.
	4, pp. 1226--1234, 2020.
	
	\bibitem{E2}
	D. Belsky et~al.,
	\newblock ``{Eleven telomere, epigenetic clock, and biomarker-composite
		quantifications of biological aging: do they measure the same thing?},''
	\newblock {\em American Journal of Epidemiology}, vol. 187, no. 6, pp.
	1220--1230, 2017.
	
	\bibitem{E3}
	S.~A. Rahman and D. Adjeroh,
	\newblock ``{Deep learning using convolutional LSTM estimates biological age
		from physical activity},''
	\newblock {\em Scientific Reports}, vol. 9, pp. 1--15, 2019.
	
	\bibitem{E8}
	J. Cole et~al.,
	\newblock ``{Brain age predicts mortality},''
	\newblock {\em Molecular Psychiatry}, vol. 23, pp. 1385--1392, 2017.
	
	\bibitem{E9}
	M. Levine,
	\newblock ``{Modeling the rate of senescence: can estimated biological age
		predict mortality more accurately than chronological age?},''
	\newblock {\em The Journals of Gerontology Series A: Biological Sciences and
		Medical Sciences}, vol. 68, 2012.
	
	\bibitem{E10}
	Z. Liu et~al.,
	\newblock ``A new aging measure captures morbidity and mortality risk across
	diverse subpopulations from {NHANES IV}: A cohort study,''
	\newblock {\em PLOS Medicine}, vol. 15, 2018.
	
	\bibitem{E11}
	I.~Cho, K. Park, and C. Lim,
	\newblock ``{An empirical comparative study on biological age estimation
		algorithms with an application of Work Ability Index (WAI)},''
	\newblock {\em Mechanisms of ageing and development}, vol. 131, pp. 69--78,
	2009.
	
	\bibitem{E12}
	A. Mitnitski, S. Howlett, and K. Rockwood,
	\newblock ``{Heterogeneity of human aging and its assessment},''
	\newblock {\em The Journals of Gerontology Series A: Biological Sciences and
		Medical Sciences}, vol. 72, pp. 877--884, 2016.
	
	\bibitem{RN14}
	A.~M.~Mughal, N.~Hassan, and A.~Ahmed,
	\newblock ``{Bone age assessment methods: A critical review},''
	\newblock {\em Pakistan Journal of Medical Sciences}, vol. 30, no. 1, pp.
	211--215, 2014.
	
	\bibitem{Final1}
	E. Tomei et~al.,
	\newblock ``{Value of MRI of the hand and the wrist in evaluation of bone age:
		Preliminary results},''
	\newblock {\em Journal of Magnetic Resonance Imaging}, vol. 39, no. 5, pp.
	1198--1205, 2014.
	
	\bibitem{x17}
	B. Neumayer et~al.,
	\newblock ``Reducing acquisition time for {MRI-based} forensic age
	estimation,''
	\newblock {\em Scientific Reports}, vol. 8, 2018.
	
	\bibitem{x18-3}
	D.~{Štern} et~al.,
	\newblock ``Automatic age estimation and majority age classification from
	multi-factorial mri data,''
	\newblock {\em IEEE Journal of Biomedical and Health Informatics}, vol. 23, no.
	4, pp. 1392--1403, 2019. 
	
	\bibitem{RN15}
	J.~H. Cole,
	\newblock ``{Neuroimaging-derived brain-age: An ageing biomarker?},''
	\newblock {\em Aging}, vol. 9, no. 8, pp. 1861--1862, 2017.
	
	\bibitem{Final2}
	S.~G. Popescu et~al.,
	\newblock ``{Deep learning methods for estimating" brain age" from structural
		MRI scans},''
	\newblock in {\em Medical Imaging with Deep Learning (MIDL)}, 2018.
	
	\bibitem{E16}
	K.~Franke et~al.,
	\newblock ``{Longitudinal changes in individual BrainAGE in healthy aging, mild
		cognitive impairment, and Alzheimer's disease},''
	\newblock {\em The Journal of Gerontopsychology and Geriatric Psychiatry}, vol.
	25, pp. 235--245, 2012.
	
	\bibitem{ipa}
	K.~Armanious \emph{et~al.}, ``{ipA-MedGAN: Inpainting of arbitrary regions in medical imaging},'' in \emph{ IEEE International Conference on Image Processing (ICIP)}, 2020, pp. 3005-3009.
	
	\bibitem{x24}
	K.~A. Bhawar and N.~K. Bhil,
	\newblock ``{Brain tumor classification using neural network based methods},''
	\newblock {\em {International Journal of Engineering Sciences \& Research
			Technology}}, vol. 5, no. 6, pp. 721--727, 2016.
	
	\bibitem{Extra2}
	K.~Armanious et~al.,
	\newblock ``{Independent brain F-FDG PET attenuation correction using a deep
		learning approach with generative adversarial networks},''
	\newblock {\em Hellenic journal of nuclear medicine}, vol. 22, no. 3, pp.
	179--186, 2019.
	
	\bibitem{x25}
	O. Ronneberger, P. Fischer, and T. Brox,
	\newblock ``U-net: Convolutional networks for biomedical image segmentation,''
	\newblock in {\em Medical Image Computing and Computer-Assisted Intervention
		(MICCAI)}, 2015, pp. 234--241.
	
	\bibitem{Final6}
	K.~Armanious et~al.,
	\newblock ``{Independent attenuation correction of whole body [18F]FDG-PET
		using a deep learning approach with generative adversarial networks},''
	\newblock {\em EJNMMI Research}, vol. 10, 2020.
	
	\bibitem{rad0}
	S.~{Abdulatif} et~al.,
	\newblock ``Towards adversarial denoising of radar micro-Doppler signatures,''
	\newblock in {\em International Radar Conference}, 2019.
	
	\bibitem{rad2}
	K.~{Armanious} et~al.,
	\newblock ``An adversarial super-resolution remedy for radar design
	trade-offs,''
	\newblock in {\em 27th European Signal Processing Conference (EUSIPCO)}, 2019.
	
	\bibitem{x11}
	T.~{Huang} et~al.,
	\newblock ``Age estimation from brain {MRI} images using deep learning,''
	\newblock in {\em IEEE 14th International Symposium on Biomedical Imaging
		(ISBI)}, 2017, pp. 849--852.
	
	\bibitem{x27}
	J.~H. Cole et~al.,
	\newblock ``Predicting brain age with deep learning from raw imaging data
	results in a reliable and heritable biomarker,''
	\newblock {\em NeuroImage}, vol. 163, pp. 115--124, 2016.
	
	\bibitem{x15}
	M. Urschler, S. Grassegger, and D. {\v{S}}tern,
	\newblock ``What automated age estimation of hand and wrist {MRI} data tells us
	about skeletal maturation in male adolescents,''
	\newblock {\em Annals of Human Biology}, vol. 42, no. 4, pp. 358--367, 2015.
	
	\bibitem{E13}
	S.~A. Rahman et~al.,
	\newblock ``Deep learning for biological age estimation,''
	\newblock {\em Briefings in bioinformatics}, 2020.
	
	\bibitem{E14}
	E. Bobrov et~al.,
	\newblock ``{PhotoAgeClock: Deep learning algorithms for development of
		noninvasive visual biomarkers of aging},''
	\newblock {\em Aging}, vol. 10, no. 11, pp. 3249--3259, 2018.
	
	\bibitem{UN1}
	R.~Tanno et~al.,
	\newblock ``{Uncertainty modelling in deep learning for safer Neuroimage
		enhancement: Demonstration in diffusion MRI},''
	\newblock {\em NeuroImage}, 2020.
	
	\bibitem{UN2}
	L.~Herzog et~al.,
	\newblock ``{Integrating uncertainty in deep neural networks for MRI based
		stroke analysis},''
	\newblock {\em Medical Image Analysis}, vol. 65, 2020.
	
	\bibitem{UN3}
	V.~{Edupuganti} et~al.,
	\newblock ``{Uncertainty quantification in deep MRI reconstruction},''
	\newblock {\em IEEE Transactions on Medical Imaging}, 2020.
	
	\bibitem{Final7}
	Q.~V. Le, A.~J. Smola, and S. Canu,
	\newblock ``Heteroscedastic Gaussian process regression,''
	\newblock in {\em International Conference on Machine Learning (ICML)}, 2005,
	p. 489–496.
	
	\bibitem{Final5}
	A.~Kendall and Y.~Gal,
	\newblock ``What uncertainties do we need in Bayesian deep learning for
	computer vision?,''
	\newblock in {\em Advances in Neural Information Processing Systems (NeurIPS)},
	2017, pp. 5580--5590.
	
	\bibitem{Extra}
	K.~Armanious \emph{et~al.}, ``{Organ-based chronological age estimation based
	on 3D MRI scans},'' in \emph{28th European Signal Processing Conference
	(EUSIPCO)}, 2020, pp. 1225--1228.
	
	
	\bibitem{x5}
	J.~Carreira and A.~Zisserman,
	\newblock ``Quo vadis, action recognition? a new model and the kinetics
	dataset,''
	\newblock {\em IEEE Conference on Computer Vision and Pattern Recognition
		(CVPR)}, pp. 4724--4733, 2017.
	
	\bibitem{x4}
	F.~N. Iandola et~al.,
	\newblock ``{SqueezeNet: AlexNet-level accuracy with 50x fewer parameters and
		{\textless}1MB model size},'' \url{http://arxiv.org/abs/1602.07360}, 2016,
	\newblock arXiv preprint.
	
	\bibitem{Final4}
	C.~Blundell et~al.,
	\newblock ``Weight uncertainty in neural networks,''
	\newblock in {\em International Conference on Machine Learning (ICML)}, 2015,
	pp. 1613--1622.
	
	\bibitem{Extra6}
	P.~LaMontagne et~al.,
	\newblock ``Oasis-3: Longitudinal neuroimaging, clinical, and cognitive dataset
	for normal aging and alzheimer disease,''
	\newblock {\em {Alzheimer's \& Dementia}}, vol. 14, 2018.
	
	\bibitem{Extra7}
	K. Schmidt,
	\newblock ``Clinical dementia rating scale,''
	\newblock in {\em Encyclopedia of Quality of Life and Well-Being Research}, pp.
	957--960. 2014.
		
	\end{thebibliography}

\end{document}